\documentclass{article}

\usepackage{packages}

\usepackage{comment}
\begin{document}



\def\TOPIC{Time Series}

\def\TITLE{Long Short-term Memory RNN}

\def\GROUP{Group 8 - Time Series}

\def\AUTHORS{
    Adrian Kjærran (adriankj) \\
    Erling Stray Bugge (erlinsb) \\
    Christian Bakke Vennerød (christbv) \\
}


\begin{titlepage}

\vbox{ }

\vbox{ }

\begin{center}
\includegraphics[width=0.40\textwidth]{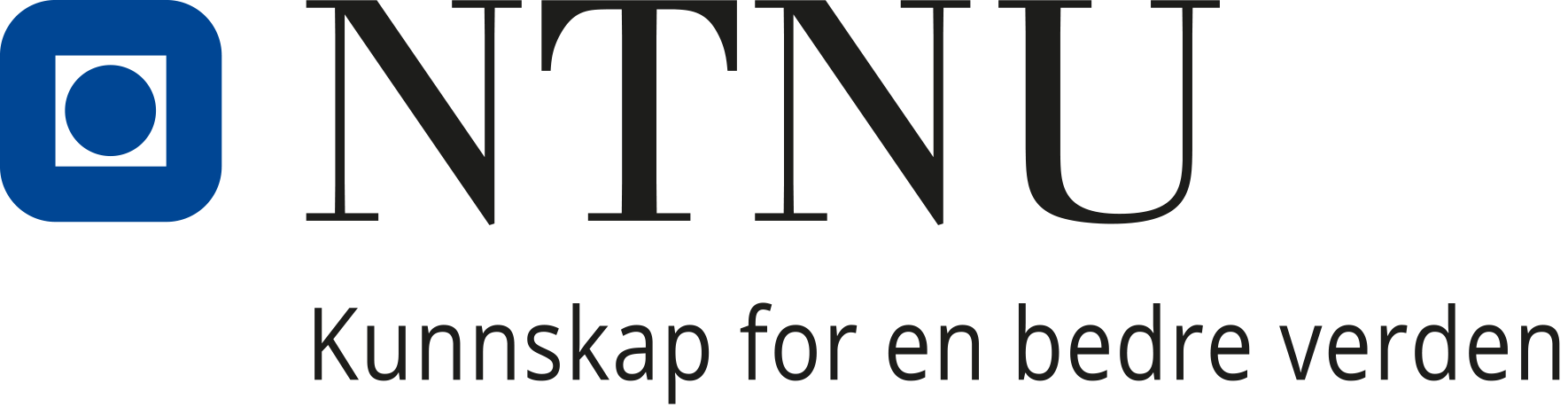}\\[1cm]
\textsc{\LARGE Department of Computer Science}\\[1.0cm]
\textsc{\Large TDT4173 - Assignment 1}\\[0.5cm]

\def\checktitle{Paper Title}
\ifx\TITLE\checktitle
    \todo[inline]{
        Remember to fill in topic name, paper title, group name, and members in "title.tex"
    }
\fi

\vbox{ }
\HRule \\[0.4cm]
{ \large \TOPIC } \\[0.3cm]
{ \huge \bfseries \TITLE}\\[0.4cm]
\HRule \\[1.5cm]
\large

\emph{Group:}\\
\GROUP

\emph{Authors:}\\
\AUTHORS

\vfill
{\large \today}
\end{center}
\end{titlepage}






\frontmatter

\begin{abstract}
    \noindent This paper is the first assignment in the course Machine Learning TDT4173, fall 2020. The project was conducted by reading through some of the latest developments within time-series forecasting methods in the scientific community over the past 5 years. The result is a summary of the essential aspects. 

In this paper, we introduce an LSTM cell’s architecture and explain how different components go together to alter the cell’s memory and predict the output. Also, the paper provides the necessary formulas and foundations to calculate a forward iteration through an LSTM. Then, the paper refers to applications and research that emphasize the strengths and opportunities of LSTMs, shown within the time-series domain and the natural language processing (NLP) domain. Finally, alternative statistical methods for time series predictions are highlighted, where the paper look into ARIMA and exponential smoothing. Nevertheless, as LSTMs can be viewed as a complex architecture, the paper assumes that the reader has a general knowledge of machine learning aspects, such as neural networks, activation functions, overfitting, underfitting, metrics and bias.

\end{abstract}
\clearpage

\tableofcontents

\listoffigures
\listoftables

\mainmatter

\section{Introduction}
\label{sec:introduction}


A time series is a sequence of historical measurements of an observable variable at equal time intervals. Time series data often exhibit patterns such as trend, seasonality and cycles: A time series has a trend if there is a long-term increase or decrease in the data, it is seasonal if the data is affected seasonally at a known and fixed frequency, e.g. time of the year or time of the day, and it is cyclical if the data either rises or falls according to an unfixed frequency, e.g. how the economy (GDP) oscillates around the long-term growth trend \citep{Nau}. Such patterns, as described above, can be exploited by a forecasting-model to infer dependencies between the past and the present, to predict the future \citep{Hyndman}.

Over the past 50+ years, many statistical, structured models have yielded the best predictions on time series data \citep{Hyndman}. Such structured models, e.g. ARIMA, try to prescribe the data generating process \citep{barker_et_al}. However, with the emergence of new machine learning models, increasingly complex data, and more computing power, the structured, statistical models are now challenged by the power of artificial intelligence \citep{bontempi_et_al}.

One subfield of artificial intelligence is machine learning, where models automatically improve through experience \citep{Goodfellow_et_al}. A class of machine learning methods designed to exhibit temporal behaviour is recurrent neural networks (RNN). An RNN is a standard feedforward neural network, but with a feedback loop, facilitating the network to handle temporal information or dynamic patterns, e.g. time series, where the value at a given time often depends on past values \citep{ALANIS}. One architecture building on the RNN architecture is LSTM, the main topic of this paper. An LSTM is an RNN, but with long short-term memory cells. The LSTM cell contains numerous internal weights and is particularly designed to capture and retain essential information from previous outputs.

\begin{figure}[h]
    \centering
    \includegraphics[scale = 0.8]{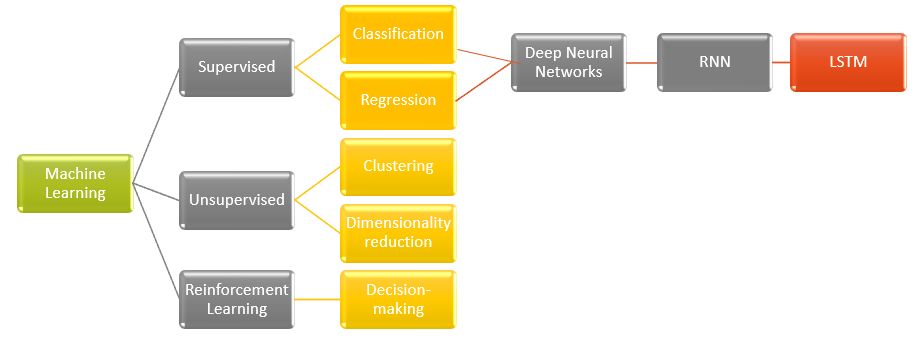}
    \caption{LSTM placed in the Machine Learning taxonomy.}
    \label{fig:tax_ai}
\end{figure}

In order for an LSTM to function properly, the weights of the internal LSTM cells must be trained. The most important aspects of the training process can be described in three steps: Firstly, the internal weights are initialized in a random configuration. Secondly, a batch of training-data is fed to the network, and an error-metric compares the output from the network and ground truth. Finally, the error is propagated back through the network, adjusting the weights within the LSTM cells through a process called back-propagation. Then this is repeated for each batch from the training data. Hence, during training, the network learns the patterns of the data by adjusting its weights towards the known correct outputs; this makes LSTM a supervised learning architecture \citep{hochreiter_et_al}. It is worth mentioning however, that LSTMs can be used on unsupervised problems too; for instance, in sequence modelling tasks such as NLP, target outputs can be generated by shifting the input one or more time-steps \citep{studass}, despite LSTM being a predominantly supervised algorithm. Figure \ref{fig:tax_ai} displays how LSTM belongs under the RNN branch in the deep-learning part of supervised learning.

\newpage

\section{Foundations}
\label{sec:foundations}

\subsection{Recurrent Neural Networks} 
LSTMs are recurrent neural networks, where the neurons are replaced by LSTMs cells. Therefore, to understand LSTMs we must understand the emergence and architecture of recurrent neural networks. The motivation behind RNNs is to tackle the problem of a standard neural network's limited memory about previous data. Standard neural network's memory is limited to their learned weight from training, and during prediction, they only utilize information about the current input $p(y_t \mid x_t)$. With exploiting knowledge of previously seen data, this becomes a difficult task. To solve this, the neurons in RNNs recurs a hidden variable $h_t$ which contains information that later iteration through the network can benefit utilize $p(y_t \mid x_t ,h_{t-1})$; hence their name recurrent neural networks \citep{zhang2020dive}. This change of architecture allows information to persist within the network and makes RNNs more suitable for predicting the outcome of a time-series compared to traditional neural networks.

\begin{figure}[h]
    \centering
    \includegraphics[width=0.7\textwidth]{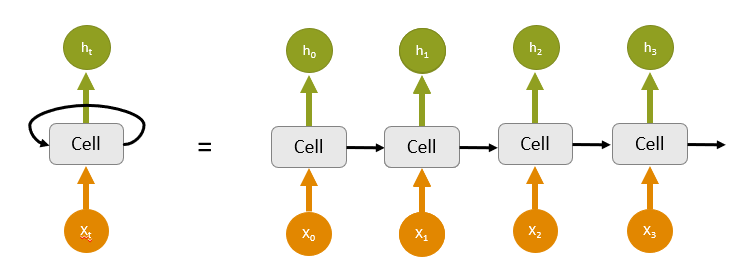}
    \caption{Unfolded representation of an RNN.}
    \label{fig:RNN_unfoldedl}
\end{figure}

\subsection{The Vanishing Gradient Problem}
In theory, the hidden state $h_t$, in each RNN cell can keep track of information for an arbitrary amount of time. However, as the hidden state $h_t$ is updated at each time step, i.e. $h_t = f(x_t, h_{t-1})$, the time dependency often causes long-term dependencies, e.g 100 time steps, to be ignored during training \citep{le2016quantifying}. Neural networks are trained through the back-propagating of an error measure through the network, and adjusting the weights of the matrices and biases such that the error is reduced for future iterations. One problem that emerges when the network becomes deeper is that the gradients tend to explode or vanish. This is because the gradients, i.e. the change of weights, are computed based on the cell's contribution to the output. The problem greatly intensifies with RNNs, since gradients are not only calculated in the "depth"-dimension, but also in the "time"-dimension \citep{Guo2013BackPropagationTT}. This problem is called the vanishing gradient problem, and it makes it difficult for standard RNNs to learn long-term dependencies in the data.

\subsection{The Idea behind LSTMs}
LSTMs were introduced to handle the vanishing gradient problem of the traditional RNNs. An LSTMs architecture is similar to an RNN, but the recurrent cells are replaced with LSTM cells \citep{hochreiter_et_al} which are constructed to retain information for extended periods. Similar to the cells in standard RNNs, the LSTM cell recurs the previous output, but also, it keeps track of an internal cell state $c_t$, which is a vector serving as long-term memory. Hence, the LSTMs have access to both short-term memory, i.e. the hidden state $h_t$, and long-term memory, i.e. the cell state $c_t$, resulting in the name: long short-term memory cells \citep{hochreiter_et_al}. The cell state can metaphorically be viewed as a conveyor belt that passes information down the line, to be used in later calculations. The content of this cell state vector is carefully altered through operations by gates inside the LSTM cell, which lets essential information and gradients to flow unchanged.

\subsection{Intuition behind LSTM} 

The main idea of the LSTM cell is to regulate the updates of the long-term memory (cell state), such that information and gradients (for training) can flow unchanged between iterations.

To give an intuition, imagine an LSTM that is to predict the next scene in a movie with two main characters, Alice and Bob. The core idea would be to let the movie scenes without Alice leave the Alice-related information completely unchanged. Through careful regulation, we could display half-an-hour of scenes with Bob without this interfering with the cell state's weights which saves information about Alice. To carefully regulate the cell-state, the LSTM uses the three yellow gates in Figure \ref{fig:LSTM_Cell}. 

\textbf{\textit{Forget gate}} – Regulates forgetting of previous information. If a scene ends, we want to forget all the information associated with the scene, but still remember information about Alice and Bob

\textbf{\textit{Input gate}} – Controls the relevance and saving of new information. When a new scene is introduced, the input gate "writes" to memory the critical information to remember about the scene.

\textbf{\textit{Output gate}} – Joins the long-term information with the short-term information to give a better prediction about the next hidden unit/output. 

These precise operations let the LSTM cell keep track of the most relevant information, update the cell state in a precise way, and choose what specific information to pay attention to when predicting new outputs.

\subsection{Components and Calculations}
\begin{figure}[h]
    \centering
    \includegraphics[width=0.6\textwidth]{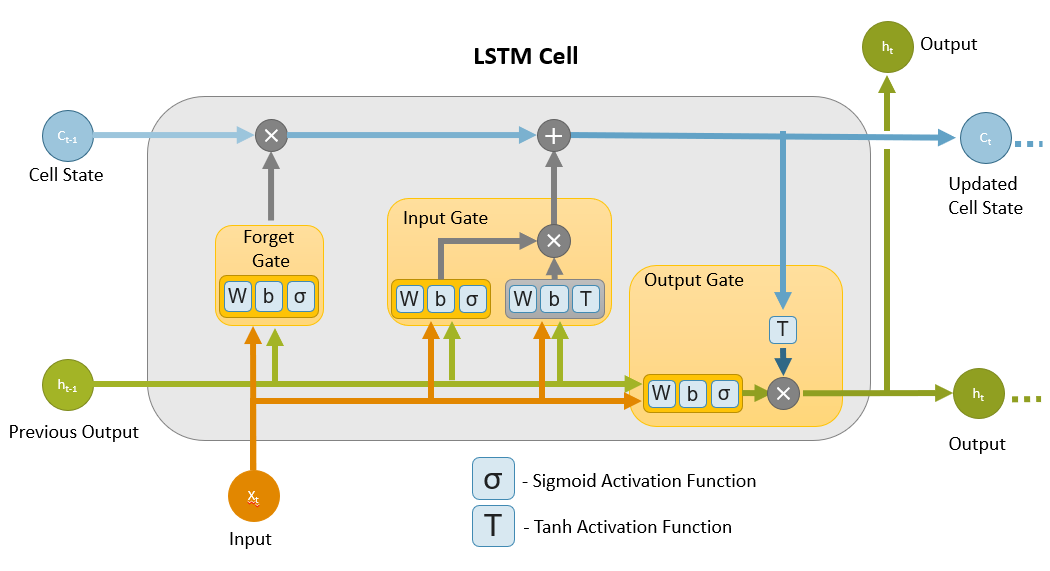}
    \caption{The Architecture of a LSTM Cell}
    \label{fig:LSTM_Cell}
\end{figure}

To understand how LSTMs work, we must thoroughly understand the components of the LSTM cell. The cell has three internal gates, namely the \textbf{forget gate}, the \textbf{input gate}, and the \textbf{output gate}, see figure \ref{fig:LSTM_Cell}. These gates control the internal cell state $c_t$, and the hidden state $h_t$. In figure \ref{fig:LSTM_Cell}, we observe in the yellow boxes how the input vector, $x_{t[n-1]}$, and the previous hidden state, $h_{t-1[m,1]}$, are multiplied respectively with four different matrices, $W_j{[m,m+n]}$, and added respectively with four different biases, $b_j{[m,1]}$. The outputs of these gates are then used to both update the cell state to form $c_{t[m-1]}$, and to calculate the new hidden state $h_{t[m,1]}$. Here, the notation within the brackets, i.e. $[m,1]$ and $[m,m+n]$, are the dimensions of the matrices. Furthermore, the "cross-sign" in the gates are a dot product, and the "plus-sign" addition.

The four combinations of a matrix $W$, a bias $b$, and an activation function, that is either the sigmoid or the tanh, can be view as a one-layer neural network inside of the cell. We define each of these one-layer neural networks as functional units, and their general formula is shown equation \ref{eq:functional_unit}.

\begin{equation}
  OutFunctionalUnit_{[m,1]} = ActivationFunction( W_{[m,m+n]}*[h_{[m,1]},x_{[n,1]}] + b_{[m,1]} )
  \label{eq:functional_unit}
\end{equation}

\textbf{Forget Gate}

The forget gate decides what information and how much information to \textbf{\textit{erase}} from the cell state. The information in the cell state $c_{t-1}$ is altered through an elementwise multiplication of the forget gate’s output $f_t$ \ref{eq:forget_gate}, see equation \ref{eq:cell_state_forget} and Figure \ref{fig:LSTM_Cell}. The forget gate’s activation function is the sigmoid function; this ensures that the output vector from the gate yields continuous values between 1 (keep) and 0 (forget). The subscript \textit{‘f’} on the matrix and bias in equation \ref{eq:forget_gate} indicates their belongingness to the forget gate.  

\begin{equation}
f_{t} = \sigma(W_{f}\cdot[h_{t-1}, x_{t}] + b_{f})
\label{eq:forget_gate}
\end{equation}
\begin{equation}
    c_f = c_{t-1} \times f_t 
    \label{eq:cell_state_forget}
\end{equation}

\textbf{Input Gate }

The next important gate is the input gate. This gate uses two functional units. The first functional unit use a tanh activation which output values between -1 and 1 and decide the \textbf{\textit{change}} of the cell state $\tilde{c}_t$. The latter use a sigmoid activation function, and it is responsible for the \textbf{\textit{magnitude}} of the change $m_{t}$. After multiplying the results of these two functional units, the input gate adds the result to the cell state, equation \ref{eq:cell_state_add}, and this completes the update of the cell state $c_t$

\begin{equation}
    \tilde{c}_{t} = tanh(W_{c}\cdot[h_{t-1}, x_{t}] + b_{c})
\end{equation}
\begin{equation}
    m_{t} = \sigma(W_{m}\cdot[h_{t-i}, x_{t}] + b_{m})
\end{equation}
\begin{equation}
    c_{t} = c_f + \tilde{c}_{t} * m_{t}
    \label{eq:cell_state_add}
\end{equation}

\textbf{Output Gate} 

The output gate uses a trained matrix $W_o$ and bias $b_o$, to fetch relevant information from the current input and previous output, equation \ref{eq:output_gate}. This information is combined with the newly adjusted cell state $c_t$ to \textbf{\textit{predict}} the next output, $h_t$, equation \ref{eq:hidden_unit}. This output recurs such that the next iteration can use it.  If there are multiple layers, this output is also used as input in the next layer, if not this output is the prediction. 

\begin{equation} 
    o_{t} = \sigma(W_{o}[h_{t-1}, x_{t}] + b_{o})
    \label{eq:output_gate}
\end{equation}
\begin{equation}
    h_{t} = o_{t} * tanh(c_{t})
    \label{eq:hidden_unit}
\end{equation}

\subsection{Pseudocode}

\begin{algorithm}[H]
\caption{One iteration through an LSTM cell}
\begin{algorithmic}
\Procedure{Iteration}{$c_{t-1}, h_{t-1}, x_t$}
\State $c_t \leftarrow$ Element-wise multiply $c_{t-1}$ with output from forget gate $f_t$ \Comment{Equation (3)}
\State $c_t \mathrel{{+}{=}} $ The output from the input gate $(\tilde{c}_t * m_t$)\Comment{Equation (6)}
\State $h_t \leftarrow$ Combination of the cell state $c_t$, input $x_t$ and previous output $h_{t-1}$ \Comment{Equations (7,8)}
\State \textbf{return} $(c_t,h_t)$
\EndProcedure
\end{algorithmic}
\end{algorithm}

\subsection{Evaluation}

To evaluate LSTMs performance, it is common to use error metrics such as root mean squared error (RMSE) and mean absolute error (MAE) \citep{mae_etc}. In general, When evaluating neural networks, it is important to evaluate the model on test data that the algorithm has not seen before. This is especially important in neural networks because their ability to learn complex structures makes them vulnerable to overfitting. In Section \ref{sec:applications} we will use these error metrics to discuss LSTMs performance.
\newpage
\section{Current Applications}
\label{sec:applications}
Due to the inherent ability of LSTMs to handle temporal data, variants of  LSTMs architectures can be used in various types of applications, in many different domains. This section focus on applications within the time-series domain and the natural language processing (NLP) domain.

\subsection{LSTMs used in forecasting solar power production}
Being able to forecast future power generation for a power plant can, among other things, help to minimize operational costs, and more efficiently plan for maintenance and improvements \cite{zuzuki_et_al}. As the power generation for solar power plants is highly dependent on the current weather situation, the resulting prediction is influenced by the performance of the weather prediction, as well as the performance of the power-generation prediction. Gensler et al. \citeyear{gensler_et_al} focused only on the last part, where they used the weather forecasts as a given input and utilized LSTMs to predict power generation.


\begin{wrapfigure}{r}{0.4\textwidth} 
    \centering
    \includegraphics[width = 0.4\textwidth]{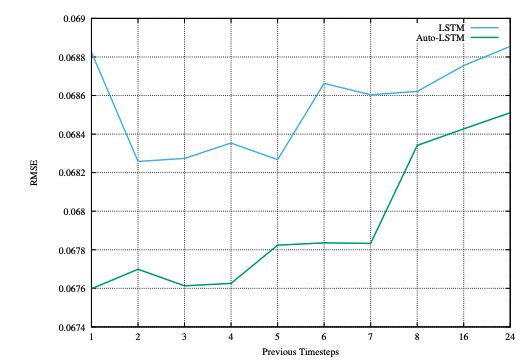}
    \caption{Choose n at the point where RMSE starts to rise \citep{gensler_et_al}.}
    \label{fig:nRMSE}
\end{wrapfigure}

\subsubsection{How LSTM was implemented}

The authors implemented both an Auto-Encoder LSTM and a traditional LSTM network. We will in this section, however, only focus on the latter one. The network consisted of four LSTM layers stacked on top of each other and an output layer with Rectified Linear Unit (ReLU) as an activation function for the predictions \citep{gensler_et_al}. The dataset was aggregated on 990 days of weather data in 3-hour resolutions from 21 different power plants. The LSTM network was trained on $n = 2$ previous time-steps to predict a new value, see figure \ref{fig:nRMSE} for an explanation of why $n = 2$ was chosen. This means that the memory cells of the LSTMs were trained by trying to predict $y_t$ based on the two past values, $\{y_{t-1}, y_{t-2}\}$, then predicting $y_{t+1}$ based on $\{y_{t-1}, y_{t-2}\}$ and so on. However, as the model's performance was based on its ability to predict 24h and 48h into the future, a recursive strategy was implemented to turn one-step predictions into a multi-step prediction. A physical forecasting model, P-PVFM, was also implemented as a reference model. The results are summarized in table \ref{tab:results_power}. We see from the Root Mean Squared Error (RMSE) and Mean Average Error (MAE) that the LSTM model prove superior to the reference model. Furthermore, whereas the reference model overestimates the power generation (positive bias), the LSTM model underestimates the power output (negative bias), although to a lower extent \citep{gensler_et_al}.

\begin{table}[h]
\centering
\begin{tabular}{l|l|ll}
          & P-PVFM & LSTM          &  \\ \cline{1-4}
Avg. RMSE & 0.1086 & 0.0724    &  \\ \cline{1-4}
Avg. MAE  & 0.0560 & 0.0368    &  \\ \cline{1-4}
Avg. BIAS & 0.0399 & -0.0073  & 
\end{tabular}
\caption{Test-results for power generation forecasting. Values closer to 0 the better.  \citep{gensler_et_al}.}
\label{tab:results_power}
\centering
\end{table}

\subsubsection{Why LSTMs works well in this scenario}
Physical models often require specific knowledge of the power plant and are more based on parameters such as solar curves, wind turbine dimensions etc. Whereas physical models can be highly accurate and offer a high degree of interpretability, they are often too dependent upon the precision of measurements. Hence, as statistical- or machine learning models do not try to explain the underlying process directly, but instead try to find a function describing the relationship between weather data (input) and power generation (output), they are more flexible and less affected by measurements errors \citep{gensler_et_al}. Furthermore, as the dataset included dates and contained a large number of features, both numerical and categorical (snowfall, rain etc.), the LSTM work even better as they can benefit from cross-series training, i.e. learn dependencies across different time-series, and thus learn not just the trend, but also seasonal patterns and level. This is also emphasized by \cite{barker_et_al}, who highlights why global, unstructured models, e.g. LSTMs, can be better than structured models, e.g. ARIMA, at capturing elements of trend, levels, and seasonality on more complex data.

\subsection{LSTMs used in Natural Language Processing}
LSTMs can be used in a variety of NLP applications, such as language modelling, where a probability distribution over words are learned. This is because the architecture of LSTMs is well suited for sequence modelling; in the same way as predicting $y_t$ based on a time series sequence $Y = {y_{t-N}, \dots, y_{t-1}}$, one can use the previous words in a sentence $W = \{w_{t-N},\dots, w_{t-1}\}$ to predict $w_t$.

\subsubsection{How LSTM was implemented}
One example of LSTMs used in NLP is with ELMo (Embeddings from Language Models) from 2018 \citep{peters_et_al}. The full details are beyond the scope of this paper. Nevertheless, the model uses bidirectional LSTMs to produce context-sensitive embeddings. Whereas language models is a probability distribution over words, an embedding is a floating-point vector-representation of words. The higher the vectors’ dimensions in the embedding, the more fine-grained relationships between words can be captured. Two key advantages of embeddings are that one can use cosine-similarity to find related words (synonyms) and use vector operations on words \citep{word2vec}. The results are shown in Table \ref{tab:results_elmo}, where SQuAD and NER are benchmarks for question answering and named entity recognition (classifying a word in a sequence to an entity), respectively. Note that newer models have surpassed the human benchmark on the SQuAD dataset \citep{squad}, and F1 is a metric which uses the harmonic weighting between precision and recall.




\begin{table}[h]
\centering
\begin{tabular}{l|l|l|ll}
          & Previous methods & Humans   & ELMo        &  \\ \cline{1-4}
SQuAD 1.1 (F1) & 84.4 & 91.2  & \textbf{87.4}  &  \\ \cline{1-4}
NER (F1) & 91.9 & 96.9 & \textbf{92.2} & 
\end{tabular}
\caption{The results of previous methods, human performance and ELMo on two different NLP benchmarks. 100 is a perfect score.}
\label{tab:results_elmo}
\centering
\end{table}

\subsubsection{Why LSTMs works well in this scenario}

One of the reasons LSTMs worked well in this scenario is their ability to be bidirectional. ELMo used two separate forward and backwards 2-layered LSTMs (bidirectional LSTM). In contrast to the forward-LSTM in ELMo, who looks at the preceding words $\{w_1, \dots, w_{k-1}\}$ in a sequence to find the probability of $w_k$, a backward-LSTM looks at the succeeding words $\{w_{k+1}, \dots, w_{N}\}$ to predict $w_k$ \citep{peters_et_al}. One advantage of this is that it enabled ELMo to learn context-sensitive embeddings \citep{peters_et_al}, meaning that a word such as "bank" will have a different vector representation if it is used in the context of a "river bank", versus in the context of "Bank of America". This is one of the reasons why ELMo could learn more patterns and relationships in the language it is trained on and thus increased its performance.

\newpage

\newpage

\section{Alternative methods}
\label{sec:improvements}
As described in the applications section, LSTM-architectures can be used in various types of forecasting applications. It is, however, important to emphasize that the type of data available and the type of problem heavily influence the right choice of method. This section covers two alternatives, statistical methods that have been widely used in the past 50+ years: ARIMA models and exponential smoothing models. After giving the foundations of these methods, we compare different strengths and weaknesses of the more statistical methods to LSTMs.

\subsection{ARIMA and its components}
ARIMA is an acronym for Autoregressive Integrated Moving Average and is a general class of linear forecasting models 
In fact, more known methods such as linear regression, random-walk and exponential smoothing are all special cases of ARIMA models \cite{Hyndman}. ARIMA models can work on both seasonal and non-seasonal data. However, we will in this section only focus on non-seasonal ARIMA models and non-seasonal data, which consist of three components: Autoregressive models (AR), Integrated (I), and Moving Average models (MA) \cite{Hyndman}.

Firstly, Autoregressive models (AR) attempt to predict a future variable, $y_t$, by creating a linear combination of past values of the variable the model aims to predict $y_{t+1}$.

\begin{equation}
    y_{t} = c + \phi_1y_{t-1} + \phi_2y_{t-2} + \dots + \phi_py_{t-p} + \epsilon_t
    \label{AutoRegressive}
\end{equation}

Here, $\epsilon_t$ is an error term, e.g. white noise, $\phi_j$ is the variables to be fitted, $j \in (1,p)$, and $c$ is the average of the period change (long-term drift). If $c$ is positive, $y_{t}$ will tend to drift upwards, whereas if $c$ is negative, $y_{t}$ will tend to drift downwards \citep{Hyndman}.

The Integrated component (I) is a component that tries to reduce the seasonality and trend in a time series. A stationary time series means that the time series has a constant mean and exhibits neither trend nor seasonality \citep{Nau}. This implies that the statistical properties are constant over time, i.e. a constant autocorrelation, meaning that the correlations between all two adjacent data points in the time series are constant. Through differencing a given number of times, a non-stationary time series can often be made stationary. Hence, $y'$ represents the observed variable differenced $d$ number of times, i.e. $(y'_t = y_t-y_{t-1})^d$.

The last component, the Moving Average (MA) model, use a linear combination of the past $q$ forecast errors to predict $y_t$.

\begin{equation}
    y_t = c + \epsilon_t + \theta_1\epsilon_{t-1} + \dots + \theta_q\epsilon_{t-q}
    \label{MovingAverage}
\end{equation}

Here, $\epsilon_t$ is an error term, and c is the average of the long-term drift. $\theta_t$ can be estimated through maximum likelihood estimation \cite{Hyndman}. 

Hence, now that we have outlined the three components making up ARIMA, we can for a given time-series difference a given number of times to make it stationary, and combine equation \eqref{AutoRegressive} and equation \eqref{MovingAverage}, to form $ARIMA(p,d,q)$, where $p$ represents the order of the autoregressive component, $d$ represents the degree of differencing, and $q$ is the order of the moving average component. That is,

\begin{equation}
    y'_t = c + \phi_1y'_{t-1} + \dots + \phi_py'_{t-p} + \theta_1\epsilon_{t-1} + \dots + \theta_q\epsilon_{t-q} + \epsilon_t
\end{equation}

\subsection{Exponential Smoothing}
Many Exponential Smoothing (ES) models are special cases of ARIMA models, and made up of an exponentially decaying autocorrelation forecast \citep{Hyndman}. Furthermore, there are also parameters for adjusting with respect to both trend and seasonality. These adjustments will be made either linearly or multiplicatively, depending on the type of trend and seasonality observed in the time series. This subsection will describe the components which make up the linear ES-model, in addition, to describe how these parameters can be used to make both a linear and a non-linear ES-model.

\subsubsection{Simple Exponential Smoothing}
The simplest form for an exponential smoothing algorithm is a weighted average of the past observations, where the weights have a decreasing magnitude as we move away in time from the event we try to predict. We use the parameter $\alpha$ to denote this decay and $h$ as a step length. As a result, $\hat{y}_{t+1|t}$ is predicted in the following manner \citep{Hyndman}:

\begin{equation}
\begin{split}
    & \textrm{Forecast Equation: } \hat{y}_{t+h|t} = l_{t} \\
    & \textrm{Smoothing Equation: } l_{t} = \alpha y_{t} + (1-\alpha)l_{t-1}
\end{split}
\end{equation}

A higher value of $\alpha$ will result in more weight on the most recent observations.

\subsubsection{Modeling trends with Exponential Smoothing}
Time series often include trends. In order to model this in the ES algorithm, a method known as Holt's linear method is used \citep{Hyndman}. There are here two components, which makes up the forecasted value, a level equation and a trend equation.

\begin{equation}\label{eq:11}
\begin{split}
    & \textrm{Forecast Equation: } \hat{y}_{t+h} = l_{t} + hb_{t}\\
    & \textrm{Level Equation: } l_{t} = \alpha y + (1-\alpha)(l_{t-1} + b_{t-1}) \\
    & \textrm{Trend Equation: } b_{t} = \beta(l_{t} - l_{t-1}) + (1-\beta)b_{t-1}
\end{split}
\end{equation}

 Now, the level equation is based on a weighted average between the last trend parameter, $b_{t-1}$,  and the last observed value of $y_{t}$. The $\beta$ will be a smoothing parameter for the trend and will have a value in the range $[0, 1]$. Depending on how much one desire to weight the new level differences as a trend or used the past calculation of the trend.
 
 The equations above now show a forecasting expanding to infinity for large values of h. Therefore, it is relevant to find methods for limiting the growth of $\hat{y}$, when h takes a significant value. This is known as damping. Mathematically, we can swap $h$ for $(\phi + \phi^{2} + ... + \phi^{h})$, and multiply $b_{t}$ with $\phi$ in the trend equation. $\phi$ will be in the range $<0,1>$.
 
\subsubsection{Modeling Seasonality with Exponential Smoothing}
For modelling seasonality with ES-models, a third smoothing parameter is required, in addition to a value to indicate the cyclical fluctuations in the seasonality. The smoothing parameter is $\gamma$, and the cycles have a duration of \textit{m} with respect to the duration of a time unit. The forecast is written \citep{Hyndman}:

\begin{equation}\label{eq:12}
\begin{split}
    & \textrm{Forecast Equation: } \hat{y}_{t+h} = l_{t} + hb_{t} + s_{t+h-m(k+1)}\\
    & \textrm{Level Equation: } l_{t} = \alpha (y_{t}-s_{t-1}) + (1-\alpha)(l_{t-1}+b_{t-1}) \\
    & \textrm{Seasonal Equation: } s_{t} = \gamma (y_{t} - l_{t-1} - b_{t-1}) + (1-\gamma)s_{t-m}
\end{split}
\end{equation}

The parameter $k$ is the integer part of $(h-1)/m$, hence, indicating the number of whole cycle lengths we forecast in the future.

\subsection{Discussion and Comparison of LSTM with Statistical Methods}

When reading papers comparing the statistical approaches to forecasting with those of deep learning, it is obvious that the results are somewhat diverging. Papers are describing that advanced machine learning models are capable of extracting highly complex relationships in the data, and utilize these to have superior forecasting compared to ARIMA and ES \citep{Siami-Namini_et_al}. On the other hand, some papers have found that the machine learning models fit the training data and perform poorly in testing, due to over-fitting \citep{Makridakis_et_al}.

First of all, the article \citep{Makridakis_et_al} concludes that an LSTM approach for forecasting single and multiple steps of a time series, based on univariate data, is a problem best solved using an ES or ARIMA model. They have found that the LSTM is capable of learning the relationships of the training data, but lacks the capability to predict new values. In other words, the LSTM will over-fit the training data. They credit the ARIMA model for restricting complexity, and therefore reducing the generalization error in the testing phase. Furthermore, the article \citep{Makridakis_et_al} discusses how machine learning algorithms need more insight into the objective of the time series to make accurate predictions.

Despite what has been described in the paragraph above, some papers strongly argue that machine learning models, such as the LSTM, are superior to statistical approaches \citep{Siami-Namini_et_al}. The research they have conducted strongly indicate that the LSTM outperforms the ARIMA model. Furthermore, they have made the study based on some of the most complex graphs available -  stock market graphs. Hence, this strongly advocates for the superiority of the LSTM. The core of their argumentation for the superiority of the LSTM is the gradient descent optimization algorithm. 

Combining the studies we have discussed so far, it is evident that there are diverging answers to whether some of the statistical approaches to time series forecasting are better or worse than an LSTM approach. It is hard to arrive at a precise conclusion. Therefore, we will discuss some of the issues we have seen in the papers, and give some guidance for choosing a time series forecasting model.

When constructing models as complex as both the ARIMA, ES and LSTM, the hyperparameters are paramount to ensure a model which can capture and communicate the relationships in the fluctuations of the time series data. Since some of the algorithms had standardized hyperparameters in their construction, it is possible to argue for the fact that they were not able to forecast at their best.

When reading the research papers, it is evident that it is not a standardized data set which has been used, but many different ones. Hence, the quantitative assessment of the different models' time series forecasting abilities is not that comparable across the research papers. 

Therefore, we can conclude that there are options to the LSTM models for time series forecasting. These models can be made linear and are in many cases great at doing time series forecasting. As a result, one should not completely disregard the more statistical processes of time series analysis, but use these as a reference and tool, since they are a lot less computationally demanding. Furthermore, we believe that as the world progresses and technology with it, we will discover a lot more complex and precise models for time series analysis, as stated in \citep{Makridakis_et_al}. This is evident in the democratization process of computer power and the availability of open-source code.

\newpage
\section{Conclusions and Further Work} 
\label{sec:conclusions}

This paper explains and presents LSTM cells and tie them to the time series domain. We have seen how LSTMs differ from traditional RNNs through their ability to deal with the vanishing gradient problem. LSTMs are also compared to more structured, traditional statistical models such as ARIMA, where we highlight the discussion around the applicability of LSTM compared to the traditional statistical approaches. Later on, different applications of the LSTM architecture are also presented, where we find that it achieves great results on predicting future power generation. Furthermore, we have seen how and why LSTMs can be applied to tasks belonging to other domains, such as NLP, and on their ability to create robust embeddings from language models.

Looking at further work, several improvements can increase the performance of LSTMs architectures. One increasingly popular implementation is deep-LSTM (DLSTM), where multiple LSTM cells are layer-wised stacked on top of each other to build more complex realizations of the data. Others are through using the concept of self-attention, outlined by \citep{waswani2017}, where LSTM architectures can be adapted to focus more on relevant parts for a given data \citep{lim2020}. Furthermore, methods such as BERT, a state-of-the-art encoder architecture released by Google in 2018, which builds upon the encoder-decoder thought from LSTM, can be used \citep{devlin2019bert}. Furthermore, many statistical, structured forecasting methods can also be used in combination with neural networks to yield great results on time series data.  

As the world becomes more data-driven than ever before, we believe that with increasing computing power, and more sharing of knowledge through the open-source community, machine learning models building on LSTM architectures will both become more widespread and accurate. We also want to emphasize that deep learning is not necessarily the best solution; classical statistical methods such as ARIMA can perform at least as good or even better depending on the dataset used or the available resources. We also want to emphasize the challenge of interpreting LSTM models. Increased depth and complexity of machine learning models can make it even harder to explain why models do as they do, which can be a severe threat under certain circumstances. To conclude, the choice of time series model depends on the data available and the data scientist’s knowledge about structures in the data. Under the right conditions, we have seen that the power of LSTMs and deep learning can result in groundbreaking results.

\newpage
\bibliography{references}

\begin{thebibliography}{22}
\providecommand{\natexlab}[1]{#1}
\providecommand{\url}[1]{\texttt{#1}}
\expandafter\ifx\csname urlstyle\endcsname\relax
  \providecommand{\doi}[1]{doi: #1}\else
  \providecommand{\doi}{doi: \begingroup \urlstyle{rm}\Url}\fi

\bibitem[Alanis and Sanchez(2017)]{ALANIS}
Alma~Y. Alanis and Edgar~N. Sanchez.
\newblock Chapter 2 - mathematical preliminaries.
\newblock In Alma~Y. Alanis and Edgar~N. Sanchez, editors, \emph{Discrete-Time
  Neural Observers}, pages 9 -- 22. Academic Press, 2017.
\newblock \doi{https://doi.org/10.1016/B978-0-12-810543-6.00002-2}.

\bibitem[Barker(2020)]{barker_et_al}
J.~Barker.
\newblock {M}achine learning in {M}{4}: {W}hat makes a good unstructured model?
\newblock \emph{International Journal of Forecasting}, 36:\penalty0 150--155,
  2020.

\bibitem[Bontempi et~al.(2013)Bontempi, Borgne, and Taieb]{bontempi_et_al}
G.~Bontempi, Y.~L. Borgne, and S.~B. Taieb.
\newblock {M}achine {L}earning {S}trategies for {T}ime {S}eries {F}orecasting.
\newblock \emph{Lecture Notes in Business Information Processing}, 2013.

\bibitem[Bouktif et~al.(2018)Bouktif, Fiaz, Ouni, and Serhani]{mae_etc}
Salah Bouktif, Ali Fiaz, Ali Ouni, and Mohamed Serhani.
\newblock Optimal deep learning lstm model for electric load forecasting using
  feature selection and genetic algorithm: Comparison with machine learning
  approaches †.
\newblock \emph{Energies}, 11:\penalty0 1636, 06 2018.
\newblock \doi{10.3390/en11071636}.

\bibitem[Devlin et~al.(2019)Devlin, Chang, Lee, and Toutanova]{devlin2019bert}
Jacob Devlin, Ming-Wei Chang, Kenton Lee, and Kristina Toutanova.
\newblock Bert: Pre-training of deep bidirectional transformers for language
  understanding, 2019.

\bibitem[Gensler et~al.(2016)Gensler, Henze, Sick, and Raabe]{gensler_et_al}
A.~Gensler, J.~Henze, B.~Sick, and N.~Raabe.
\newblock Deep {L}earning for solar power forecasting - {A}n approach using
  {A}uto{E}ncoder and {L}{S}{T}{M} {N}eural {N}etworks.
\newblock \emph{IEEE International Conference on Systems, Man and Cybernetics
  (SMC)}, 2016.

\bibitem[Goodfellow et~al.(2016)Goodfellow, Bengio, and
  Courville]{Goodfellow_et_al}
Ian Goodfellow, Yoshua Bengio, and Aaron Courville.
\newblock \emph{Deep Learning Book}.
\newblock 2016.
\newblock \url{https://deeplearningbook.org}. Accessed on 12.10.2020.

\bibitem[Guo(2013)]{Guo2013BackPropagationTT}
Jiang Guo.
\newblock Backpropagation through time.
\newblock 2013.

\bibitem[Hochreiter and Schmidhuber(1997)]{hochreiter_et_al}
Sepp Hochreiter and Jürgen Schmidhuber.
\newblock Long short-term memory.
\newblock \emph{Neural computation}, 9:\penalty0 1735--80, 12 1997.
\newblock \doi{10.1162/neco.1997.9.8.1735}.

\bibitem[Hyndman and Athanasopoulos(2018)]{Hyndman}
R.J. Hyndman and G.~Athanasopoulos.
\newblock \emph{Forecasting: principles and practice, 2nd. edition}.
\newblock 2018.
\newblock \url{https://OTexts.com/fpp2}. Accessed on 09.10.2020.

\bibitem[Le and Zuidema(2016)]{le2016quantifying}
Phong Le and Willem Zuidema.
\newblock Quantifying the vanishing gradient and long distance dependency
  problem in recursive neural networks and recursive lstms, 2016.

\bibitem[Lim and Stefan(2020)]{lim2020}
Bryan Lim and Zohren Stefan.
\newblock Time series forecasting with deep learning: A survey.
\newblock 2020.

\bibitem[Mikolov et~al.(2013)Mikolov, Sutskever, Chen, Corrado, and
  Dean]{word2vec}
Tomas Mikolov, Ilya Sutskever, Kai Chen, Greg~S Corrado, and Jeff Dean.
\newblock Distributed representations of words and phrases and their
  compositionality.
\newblock In \emph{Advances in Neural Information Processing Systems 26}, pages
  3111--3119. 2013.
\newblock URL
  \url{http://papers.nips.cc/paper/5021-distributed-representations-of-words-and-phrases-and-their-compositionality.pdf}.

\bibitem[Nau(2020)]{Nau}
Robert Nau.
\newblock Statistical forecasting: Notes on regression and time series
  analysis, 2020.

\bibitem[Nordanger(2020)]{studass}
Ole~P. Nordanger.
\newblock Blackboard. discussion forum under time-series - "contextualize
  lstm".
\newblock 2020.

\bibitem[Peters et~al.(2018)Peters, Neumann, Iyyer, Gardner, Clark, Lee, and
  Zettlemoyer]{peters_et_al}
M.~E. Peters, M.~Neumann, M.~Iyyer, M.~Gardner, C.~Clark, K.~Lee, and
  L.~Zettlemoyer.
\newblock Deep {C}ontextualized word representations.
\newblock \emph{Allen Institute for Artificial Intelligence}, 2018.

\bibitem[Rajpurkar(2020)]{squad}
P~Rajpurkar.
\newblock Stanford question answering dataset (squad) 2.0 - leaderboard, 2020.
\newblock URL \url{https://rajpurkar.github.io/SQuAD-explorer/}.

\bibitem[Sima Siami-Namini(2019)]{Siami-Namini_et_al}
Akbar Siami~Namin Sima Siami-Namini, Neda~Tavakoli.
\newblock A comparative analysis of forecasting financial time series using
  arima, lstm, and bilstm.
\newblock 2019.
\newblock \url{https://arxiv.org/pdf/1911.09512.pdf}.

\bibitem[Spyros~Makridakis(2018)]{Makridakis_et_al}
Vassilios~Assimakopoulos Spyros~Makridakis, Evangelos~Spiliotis.
\newblock Statistical and machine learning forecasting methods: Concerns and
  ways forward.
\newblock 2018.
\newblock
  \url{https://journals.plos.org/plosone/article?id=10.1371/journal.pone.0194889}.

\bibitem[Vaswani et~al.(2018)Vaswani, Shazeer, Parmar, Uszkoreit, Jones, Gomez,
  Kaiser, and Polosukhin]{waswani2017}
Ashish Vaswani, Noam Shazeer, Niki Parmar, Jakob Uszkoreit, Llion Jones,
  Aidan~N. Gomez, Lukasz Kaiser, and Illia Polosukhin.
\newblock Attention is all you need.
\newblock 2018.

\bibitem[Zhang et~al.(2020)Zhang, Lipton, Li, and Smola]{zhang2020dive}
Aston Zhang, Zachary~C. Lipton, Mu~Li, and Alexander~J. Smola.
\newblock \emph{Dive into Deep Learning}.
\newblock GitHub, 2020.
\newblock \url{https://d2l.ai}.

\bibitem[Zuzuki(2015)]{zuzuki_et_al}
A.~Zuzuki.
\newblock {F}orecaster introduction, 2015.
\newblock URL
  \url{https://www.dnvgl.com/services/forecaster-introduction-3848}.
\newblock [Online; accessed 17. september 2020].

\end{thebibliography}
\end{document}